\documentclass[journal,twoside,web]{ieeecolor}
\usepackage{tmi}
\usepackage{cite}
\usepackage{amsmath,amssymb,amsfonts}
\usepackage{graphicx}
\usepackage{textcomp}

\usepackage{url}
\usepackage{color}
\usepackage{multirow}
\usepackage{wrapfig}
\newtheorem{theorem}{Theorem}
\usepackage{algorithm}
\usepackage{algpseudocode}
\usepackage{comment} 
\usepackage{hyperref}
\AtBeginDocument{%
  \hypersetup{%
    hidelinks,
    allcolors=black%
  }%
}
\newcommand{\new}[1]{\texorpdfstring{\textcolor{black}{#1}}{\detokenize{#1}}}
\newcommand{\newnew}[1]{\textcolor{black}{#1}}
\newcommand{\newnewnew}[1]{\textcolor{black}{#1}}

\def\BibTeX{{\rm B\kern-.05em{\sc i\kern-.025em b}\kern-.08em
    T\kern-.1667em\lower.7ex\hbox{E}\kern-.125emX}}
\begin{document}
\title{SAMReg: SAM-enabled Image Registration with ROI-based Correspondence}
\author{Shiqi Huang, Tingfa Xu, Ziyi Shen, Shaheer Ullah Saeed, Wen Yan, Dean Barratt, Yipeng Hu
\thanks{This work was supported by the International Alliance for Cancer Early Detection, a partnership between Cancer Research UK [C28070/A30912; C73666/A31378], Canary Center at Stanford University, the University of Cambridge, OHSU Knight Cancer Institute, University College London and the University of Manchester. }
\thanks{S. Huang, T. Xu are with Beijing Institute of Technology (e-mail: huangsq, ciom\_xtf1@bit.edu.cn). }
\thanks{S. Huang, Z. Shen, S. Saeed, W. Yan, D. Barratt and Y. Hu are with University College London (e-mail: yipeng.hu@ucl.ac.uk).}}

\maketitle

\begin{abstract}
This paper describes a new spatial correspondence representation based on paired regions-of-interest (ROIs), for medical image registration. The distinct properties of the proposed ROI-based correspondence are discussed, in the context of potential benefits in clinical applications following image registration, compared with alternative correspondence-representing approaches, such as those based on sampled displacements and spatial transformation functions. These benefits include a clear connection between learning-based image registration and segmentation, which in turn motivates two cases of image registration approaches using (pre-)trained segmentation networks. Based on the segment anything model (SAM), a vision foundation model for segmentation, we develop a new registration algorithm SAMReg, which does not require any training (or training data), gradient-based fine-tuning or prompt engineering.
The proposed SAMReg models are evaluated across five real-world applications, including intra-subject registration tasks with cardiac MR and lung CT, challenging inter-subject registration scenarios with prostate MR and retinal imaging, and an additional evaluation with a non-clinical example with aerial image registration. The proposed methods outperform both intensity-based iterative algorithms and DDF-predicting learning-based networks across tested metrics including Dice and target registration errors on anatomical structures, and further demonstrates competitive performance compared to weakly-supervised registration approaches that rely on fully-segmented training data. Open source code and examples are available at \href{https://github.com/sqhuang0103/SAMReg.git}{SAMReg}.
\end{abstract}


\section{Introduction}
\begin{figure}[!t]
    \centering
    \includegraphics[width=0.5\textwidth]{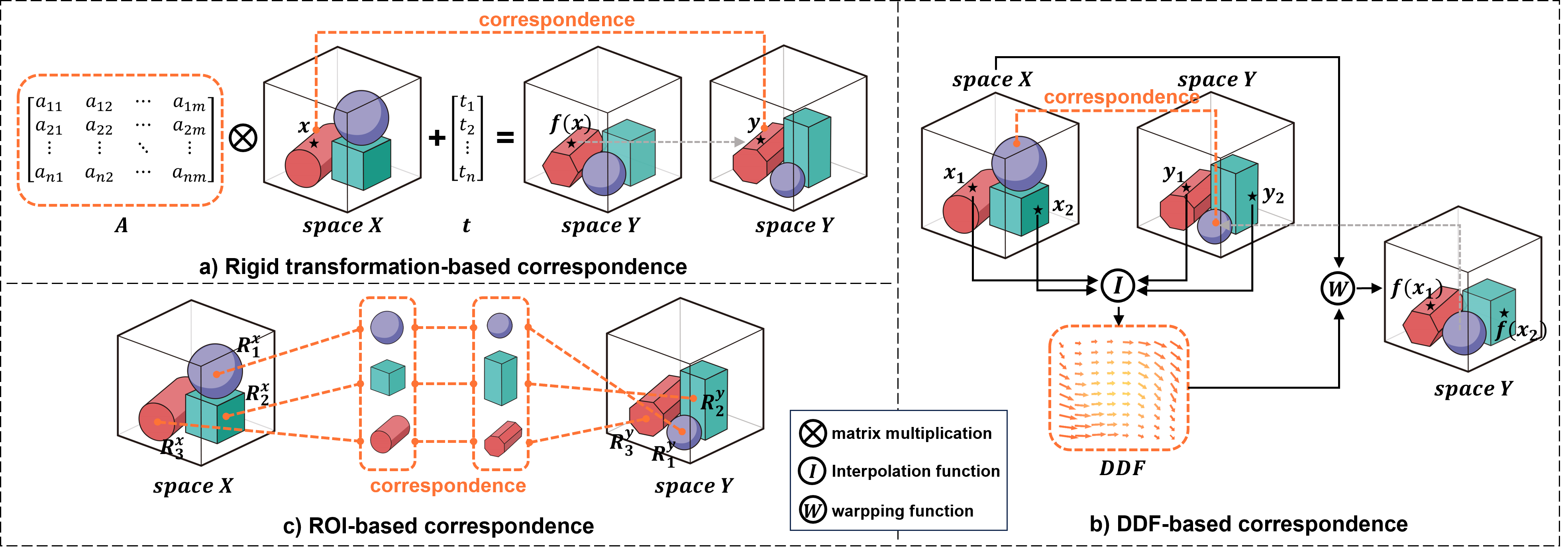}
    \caption{The representation diagram of a) rigid transformation-based, b) DDF-based and c) ROI-based correspondence. The ROI-based correspondence is exhibited in a concise and elegant manner.}
    \vspace{-2em}
    \label{fig:motivate}
\end{figure}

Both segmentation and registration are fundamental tasks in medical image analysis, enabling a wide range of clinical applications. Traditionally, a segmentation task takes one input image and searches for one or more \new{regions of interest} (ROIs). The segmentation is often represented by binary mask(s) or forms of ROI boundaries (e.g. point sets~\cite{amini1998coupled,brigger2000b} and parametric splines~\cite{li1995contour,zhu2008contour}), whilst pairwise registration takes a pair of images as input and outputs a spatial \newnew{transform that aligns one to another}. Such spatial transformation can be represented by dense displacement fields (DDFs) or other parametric functions (e.g. rigid, affine and control-point-based splines). 

Modern deep learning-based approaches adopted the same paradigms to learn the predictive relationships between respective input and output. Formulating segmentation and registration as machine learning tasks, recent works have increasingly exploited the inter-relationships between the two.
Using multi-task learning, for example, segmentation and registration models can be jointly optimised, as two weight-sharing models (e.g.~\cite{xu2019deepatlas,kang2022dual}). The two networks can also be chained sequentially. Some used segmented ROIs as weak labels to supervise or regularise registration network training (e.g.~\cite{hu2018weakly,kim2021cyclemorph}). 
Before deep-learning was adopted in segmentation, one of the most effective segmentation methods was based on registration, which propagates segmented ROIs from segmented reference images to images-to-segment~\cite{collins1997animal, zhuang2010registration}. 
The label-propagation-then-fusion idea has motivated another class of joint learning methods between the two, in which the registration was used to provide pseudo or intermediate segmentation labels, for supervising segmentation networks~\cite{hu2018label,liu2023contrastive}. What was commonly observed from the above-discussed work is the benefits of combining segmentation and registration, although they remain to be considered as two distinct tasks, with different forms of output, loss functions and training strategies. One of the characteristic differences, for example, is that registration often requires spatial resampling for computing unsupervised and/or weakly-supervised losses~\cite{detone2016deep}, whilst segmentation does not.

Previous work argued that, for a type of registration-enabled applications, predicting the ROIs on the fixed images given the ROI-labelled moving images is the practical aim of such registration, without requiring estimating unnecessary dense correspondence (those ``outside'' of the ROIs) or pixel-level correspondence~\cite{hu2019conditional}. As a result, no spatial resampling is needed in the proposed conditional segmentation algorithms. Our preliminary results summarised in a conference paper proposed a more general correspondence representation, completely based on ROI pairs, without point-to-point correspondence or those represented by parametric functions.
In this work, we aim to further explore the synergy between the underlying aims of segmentation and registration, localising ROIs and correspondence, respectively.

Studying such connections between segmentation and registration facilitate many interesting discussions in the field of medical image analysis.
First, we discuss in greater depth the proposed ROI-level correspondence representation, highlighting its generality and efficiency. This new representation is applicable to most if not all existing registration applications and, in its limiting cases, equivalent to DDF-predicting algorithms. It is also efficient in its versatility which allows only regions-of-registration-interest to be focused on and optimised. Further, we discuss two different approaches to estimate correspondence-representing ROI pairs and their respective probabilistic interpretations.
Second, taking advantage of the recent development of vision foundation models for semantic segmentation, such as segment anything models (SAMs), it is computationally efficient to develop our proposed registration algorithms with very little to none training. Therefore, the resulting algorithms, named ``SAMReg'' does not require costly clinical data, with or without labels.

\section{Related work}

\subsection{SAM-based applications other than segmentation}
Unlike traditional segmentation models that require extensive training data for specific tasks, SAM~\cite{kirillov2023segment} is designed to be highly adaptable in natural image segmentation and can perform various segmentation tasks with minimal or no additional training. Its state-of-the-art performance on various benchmarks has led to its application in fields such as medical imaging~\cite{ma2024segment,cheng2023sammed2d}, autonomous driving~\cite{cheng2023segment}, and remote sensing~\cite{wang2024samrs} segmentation. \new{Notably, recent developments have rapidly advanced its utility in automated medical segmentation, such as utilizing multi-atlas pseudo prompts for clinical analysis~\cite{fan2025ma} and multi-modal inputs for prompt-free specific cancer targeting~\cite{zhang2025generalist}.} Beyond its role in segmentation, SAM functions as a 2-dimensional dense prediction tool, offering broader applications in various computer vision tasks.

One notable application is its ability to extend 2D clusters into additional dimensions. For instance, by incorporating a third spatial coordinate, SAM enriches the semantic understanding of each 2D slice, enhancing tasks such as depth estimation~\cite{shvets2024joint,oquab2023dinov2} and 3D reconstruction~\cite{chen2024meshanything,cen2023segment}. Similarly, when applied to the temporal dimension, SAM’s ability to capture regions of interest (ROIs) across sequential frames aligns with the requirements of object tracking~\cite{yang2023track,wang2023tracking}.
Another key application lies in adaptive local processing. SAM can be used to selectively manipulate certain objects in image synthesis or to enhance local image quality, offering precise and context-aware improvements within images~\cite{fei2024lightening,chen2024harmonization}. 
In this paper, SAM is applied to \new{training-free} image registration tasks, where two or more images are taken as input, and the output is a correspondence rather than semantic segmentation. \new{Specifically, rather than isolating predefined semantic boundaries, we leverage SAM's unbiased and dense feature extraction to capture any structural context. These spatial partitions act as robust local anchors, guiding the cross-image alignment.}

\subsection{Learning-based image registration}
Both rigid and deformable image registration seek to establish correspondences between two or more images to achieve alignment. These correspondences are typically represented using one of two approaches: parametric transformations or Dense Displacement Fields (DDFs)~\cite{detone2016deep}. Parametric functions, such as rigid, affine, and spline-based transformations~\cite{rohr2001landmark}, describe the mapping of specific locations between images. DDFs, as a more general representation, define the transformation for each paired sample and are often learned through learning-based registration algorithms~\cite{detone2016deep}.

Early learning-based registration methods utilized supervised learning, where the deformation field was obtained either manually or from classical approaches~\cite{rohe2017svf}. VoxelMorph was among the first to introduce unsupervised learning for brain MRI registration~\cite{balakrishnan2019voxelmorph}, while LabelReg pioneered weakly-supervised learning for high-fidelity alignment of anatomical landmarks~\cite{hu2018weakly}. Subsequent research expanded these paradigms, exploring diverse network architectures~\cite{krebs2017robust}, loss functions~\cite{chen2022transmorph,mok2022affine}, and transformations incorporating inverse consistency or symmetric constraints~\cite{kim2021cyclemorph,tian2023gradicon,qiu2021learning}. However, these learning-based methods faced challenges with hyperparameter tuning, requiring retraining for each new regularization setting, which limited their flexibility. This limitation led to the development of foundation models like UniGradICON~\cite{tian2024unigradicon}, which is trained with a fixed set of hyperparameters across multiple datasets and can handle both in-distribution and out-of-distribution tasks without retraining. However, UniGradICON requires heavy pre-training efforts and specific fine-tuning when applied to unseen domains.

Recent efforts in either utilising large foundation models~\cite{huang2024one, song2024dino,shi2024path} and developing registration-specific foundation models~\cite{tian2023gradicon,wu2024diff} have been reported. \new{The majority of existing work aims to predict inter-image correspondence represented by pixel-level displacements or parametric spatial transformation functions}, thus requiring spatial resampling during training or fine-tuning to compute loss function values, a distinct difference from the work presented in this paper.

\subsection{Our preliminary conference presentation}
This work extends our preliminary results presented at MICCAI 2024. 
In this extended version, we adopt a novel perspective in introducing the representation and provide a more detailed algorithmic description, such as the adaptation details from 2D SAM to 3D registration with pseudo code, as well as newly added discussion and derivation on probabilistic interpretation of the segmentation-registration connection. In addition, the evaluation has been expanded to include five datasets, including both 2D and 3D datasets across various modalities such as RGB, MRI, and CT. All ablative quantitative and qualitative results have been updated and are exclusive to the current study. Moreover, a new discussion section has been included to share our experience in designing and experimenting with the new correspondence representation and its applications.
In summary, our main contributions are as follows:
\begin{enumerate}
    \item We pioneer a new correspondence representation for registration using a set of corresponding paired ROIs.
    \item We introduce the general-purpose SAMReg by segmenting corresponding ROIs from both images, without requiring any labels or fine-tuning. To our knowledge, it is the first application of SAM in image registration. Code and demo are available at \href{https://github.com/sqhuang0103/SAMReg.git}{SAMReg}.
    \item Experimental results on multiple real clinical datasets show that our methods outperform the widely adopted registration tools and unsupervised registration algorithms, and are competitive with weakly-supervised registration approaches.
\end{enumerate}

\begin{figure*}[!t]
    \centering
    \includegraphics[width=0.8\textwidth]{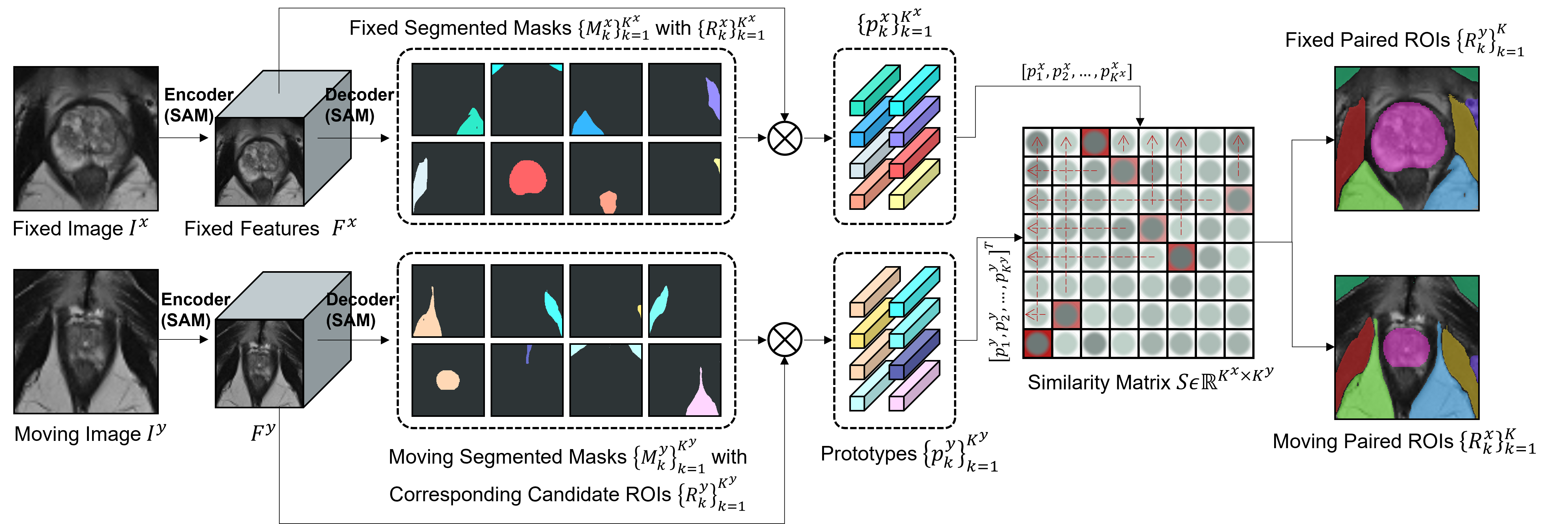}
    \caption{The architecture of the proposed SAM-Reg algorithm, \new{designed as a sequential and decoupled pipeline.}
    The process can be divided into two primary steps: ROI embedding and ROI matching. \new{1) ROI embedding: The SAM-based module functions as an independent, static feature and mask extractor.} Initially, fixed and moving images $I^x$ and $I^y$ are encoded into feature representations $F^x$ and $F^y$, which are then decoded to produce segmented masks $M_k^x$ and $M_k^y$ highlighting various ROIs $R^x_k$ and $R^y_k$. These ROIs are embedded into prototype vectors $[p^x_1,p^x_2,...,p^x_{K^x}]$ and $[p^y_1,p^y_2,...,p^y_{K^y}]$, respectively. \new{2) ROI matching: The registration module acts as a subsequent processor utilizing these static outputs.} It employs a similarity matrix $S \in \mathbb{R}^{K^x \times K^y}$ to identify and match the most similar candidate ROIs from the moving image to each fixed ROI. \new{Optionally, these established correspondences can serve as robust spatial anchors to guide the deformation field optimization}}
    \label{fig:pipeline}
\end{figure*}

\section{Method}
\subsection{One Registration is Worth Two Segmentations: ROI-Based Correspondence Modeling}
\label{sec:roi-correspondence}

\subsubsection{A brief perspective on ROI-based correspondence}
\label{ROI}
For a given class $C_k$, a segmentation task is to estimate the binary class probability vector $\{\hat{c}_n\}_{n=1}^{N}$ over $N$ voxel locations, thereby forming a segmented region of interest (ROI) $\hat{R}_k$. When applied to two images $I^x$ and $I^y$, the segmentator identifies the ROIs $R^x_k$ and $R^y_k$, of the same class $C_k$, in each image. These two ROIs, representing the same segmentation class, are considered corresponding, thereby locally registering $I^x$ and $I^y$ - local to the ROIs.

In multi-class segmentation, which estimates joint probability of a random vector $\textbf{C}=[C_1,...,C_K]^\top\in\mathbb{R}^{K}$ for $K$-class probabilities, the corresponding segmented ROI sets $\{R^{x}_k\}_{k=1}^{K}$ and $\{R^{y}_k\}_{k=1}^{K}$, derived by binarizing the estimated class probability vectors $\{\hat{\textbf{c}}_n\}^{N}_{n=1}$ with $\hat{\textbf{c}}_n^x \sim \textbf{C}$, exhibit a more comprehensive correspondence between $I^x$ and $I^y$. It is a sufficiently comprehensive representation form for both local and global correspondence, with ROIs with smaller and larger areas/volumes, respectively, as well as for both sparse and dense correspondence, determined by the spatial distribution of these $K$ pairs of ROIs. \newnew{In this section, we formalize such an ROI-based correspondence representation and discuss registering $I^x$ and $I^y$ by estimating this correspondence.}

\subsubsection{ROI pairs as a sufficient representation}

\begin{theorem}
Denoting $\textbf{x}$ and $\textbf{y}$ as spatial locations at respective moving- and fixed image spaces (\textit{i.e.} coordinate systems), it is sufficient for $K$ pairs of regions-of-interest (ROIs), denoted as $\{(R^{x}_k, R^{y}_k)\}_{k=1}^{K}$, to indicate any spatial correspondence between the two image spaces, if $K$ is sufficiently large, where $R^{x}_k = \{\textbf{x}_l\}_{l=1}^{L^x_k}$ and $R^{y}_k = \{\textbf{y}_l\}_{l=1}^{L^y_k}$ are two sets of $L^x_k$ and $L^y_k$ spatial locations that represent the same (corresponding) ROIs, in the moving- and fixed image spaces, respectively.
\end{theorem}

This is intuitive in an extreme case, in which a large number ($K=N$) of single-voxel (\textit{i.e.} $L^x_k=L^y_k=1$) ROIs are used, the ROI pairs can be sampled at $N$ voxels to an equivalent DDF representation $\{(R^{x}_k = \textbf{x}_k, R^{y}_k=\textbf{y}_k)\}_{k=1}^{N}$. 

Therefore, an informal proof for Theorem 1 can be understood by iterating the following process until sufficiency is reached:
When a point-to-point correspondence $\textbf{y}^*=f(\textbf{x}^*)$ that has NOT been represented by current ROI pairs $\{(R^{x}_k, R^{y}_k)\}_{k=1}^{K}$, one can always sample an additional ROI pair $\{(R^{x}_{K+1}=\textbf{x}^*, R^{y}_{K+1}=\textbf{y}^*)\}$, such that the resulting ($K+1$) ROI pairs $\{(R^{x}_k, R^{y}_k)\}_{k=1}^{K+1} = \{(R^{x}_k, R^{y}_k)\}_{k=1}^{K} \cup \{(R^{x}_{K+1}, R^{y}_{K+1})\}$ suffice.

\subsubsection{Properties of ROI-based correspondence} 
We discuss the properties of the ROI-based correspondence as follows: 
\begin{enumerate}
    \item Paired ROIs are practically convenient as binary mask pairs, a well-established data structure in many computational libraries;
    \item Paired ROIs can represent both local or global correspondence, if they are sufficiently small or large, respectively;
    \item Paired ROIs can represent both dense and sparse correspondence, dependent on the potentially pre-configurable spatial distribution of the ROI locations;
    \item If dense-level correspondence is required, 
    the correspondence at any spatial location can be inferred. We discuss specific algorithms in Sec.~\ref{sec:align};
    \item If dense-level correspondence is not required, omitting smoothness regularization (\textit{e.g.}, rigidity or bending energy) during estimation may improve registration efficiency with ROI-based representation;
    \item Paired ROIs are 
    independent of 
    image resolutions;
    \item Overlapping ROIs enable one-to-many correspondence;
    \item Paired ROIs have \new{the potential} to readily represent uncertainty in correspondence, by using overlapping ROIs in both images or by not using any.
\end{enumerate}

\newnew{These properties motivate treating registration as an ROI-pair estimation problem. We therefore introduce a probabilistic formulation for predicting corresponding ROI pairs from two images.}

\subsubsection{A probabilistic formulation for ROI-pair estimation}
\label{One registration}
In this section, we consider two cases to search \new{for} corresponding ROI pairs $\{R^x_k,R^y_k\}^K_{k=1}$, and their probabilistic links to \newnew{the multi-class segmentation results separately estimated from $I^x$ and $I^y$.}

\paragraph{Case 1: Segmenting the Same ROIs from Two Images} 
In this case, the segmentation algorithm (\textit{e.g.}, one or two neural networks) is capable of segmenting the same classes $\textbf{C}$ of ROIs in both images $I^x$ and $I^y$. Following notations in Sec.~\ref{ROI}, \new{obtaining ROI pairs amounts to estimating the joint probability given two images $\prod_{n=1}^{N} p_n(\textbf{C} | I^{x}, I^{y})$,} \newnew{under the conditional independence assumption given the shared ROI class variable $\textbf{C}$,} where
\begin{equation}
\begin{split}
    p_n(\textbf{C} | I^{x},I^{y}) & = \frac{p_n(I^x,I^y | \textbf{C})\cdot p_n(\textbf{C})}{p_n(I^x,I^y)}\\
    & = \frac{p_n(I^x | \textbf{C}) \cdot p_n(I^y|\textbf{C}) \cdot p_n(\textbf{C})}{\sum_{\textbf{C}}p_n(I^x,I^y | \textbf{C}) \cdot p_n(\textbf{C})}\\
    &\propto p_n(\textbf{C} | I^{x}) \cdot p_n(\textbf{C} | I^{y}),
\end{split}
\end{equation}
which is subject to normalization constants.
This approach typically requires predefined ROI types and trained segmentation networks. Notably, these segmented training datasets are exactly the same as those required for training weakly-supervised registration algorithms~\cite{hu2018weakly,hu2019conditional}.

\paragraph{Case 2: Matching the Segmented ROI Candidates} 
Assume an alternative ``unsupervised'' segmentation can be used to obtain two independent sets of candidate ROIs. Specifically, for image $I^x$, a random vector $\textbf{C}^x=[C^x_1,...,C^x_{K^x}]^\top\in\mathbb{R}^{K^x}$ representing $K^x$-class probabilities is estimated, and the segmented ROI set $\{R^{x}_k\}_{k=1}^{K^x}$ is derived by thresholding the estimated class probability vectors $\{\hat{\textbf{c}}_n^x\}^{N}_{n=1}$, where $\hat{\textbf{c}}_n^x \sim \textbf{C}^x$. A similar process is applied to image $I^y$, yielding ROIs $\{R^{y}_k\}_{k=1}^{K^y}$ via estimating $K^y$-class probabilities ${\hat{\textbf{c}}_n^y}$, where $\hat{\textbf{c}}_n^y$ is the class probabilities for $K^y$ classes. To identify corresponding ROI pairs, we search for the subset of common ROI classes $\textbf{C}$, which corresponds to the joint probability between the two conditionally independent ``candidate probabilities'', \textit{i.e.},
\begin{equation}
\begin{split}
        p_n(\textbf{C} | I^{x},I^{y}) &= p_n(\textbf{C}^x,\textbf{C}^y | I^{x},I^{y})\\ &= p_n(\textbf{C}^x | I^{x},I^{y}) \cdot p_n(\textbf{C}^y | I^{x},I^{y}) \\ &\propto p_n(\textbf{C}^x | I^{x}) \cdot p_n(\textbf{C}^y | I^{y}).
\end{split}
\end{equation}
In the following Sec.~\ref{sec:seg_match}, we propose a specific algorithm using this strategy.

\begin{table*}
\centering
\caption{\new{Summary of the datasets used for evaluating the proposed SAMReg algorithm.}}
\label{tab:dataset}
\footnotesize
\setlength{\tabcolsep}{1.7pt}
 \renewcommand{\arraystretch}{1.2} 
\begin{tabular}{c|c|c|c|c|c} 
\hline
\textbf{Dataset} & \textbf{Modality} & \textbf{Dimension} & \textbf{Task Type} & \textbf{Quantity} & \textbf{Institution} \\ 
\hline
\textbf{Prostate~} & T2-weighted MR & 3D & Inter-subject & 878 cases & University College London Hospital \\
\textbf{Cardiac} & Cine MR & 3D & Intra-subject & 150 pairs & University Hospital of Dijon \\
\textbf{Lung} & FBCT & 3D & Intra-subject & 20 pairs & Radboud University Medical Center \\
\hline
\textbf{Retina} & Retinal imaging & 2D & Intra-subject & 134 pairs & Papageorgiou Hospital \\
\textbf{Aerial~} & QuickBird-acquired & 2D & Non-clinical & 1734 pairs & City of Zurich \\
\hline
\end{tabular}
\end{table*}

\subsection{Segment Everything and Match: A Training-Free Registration Algorithm}
\label{sec:seg_match}

\subsubsection{The SAMReg algorithm}
\label{sec:align}
\newnew{Building upon the probabilistic ROI correspondence formulation introduced in Case 2 of Sec.~\ref{One registration}, we propose the SAMReg algorithm in this section. Since the multiple ROIs generated by SAM segmentation lack explicit ``one-to-one'' correspondence, SAMReg is designed as a two-stage pipeline consisting of ROI embedding followed by ROI matching, as illustrated in Fig.~\ref{fig:pipeline}.}

\textbf{\textit{\newnew{a) ROI Embedding:} }}
Leveraging the image encoder $\mathcal{E}$ in SAM, we embed the 2D input images $I\in \mathbb{R}^{H\times W}$ into features $F\in \mathbb{R}^{H^\prime \times W^\prime \times C^\prime}$: 
\begin{equation}
\label{eq:encode}
    F^x = \mathcal{E}(I^x), \quad F^y = \mathcal{E}(I^y).
\end{equation}
By setting the SAM decoder $\mathcal{D}$ to the \textit{everything} mode with a simple outlier-removing filter (further details in Sec.~\ref{sec:exp_results}), we exploit its generality to produce a comprehensive array of segmentation masks for each image, obtaining mask sets $\{M_k\}_{k=1}^{K}$, representing ROIs $\{R_k\}_{k=1}^{K}$:
\begin{equation}
\label{eq:decode}
    \{M_k\}_{k=1}^{K} = \mathcal{D}(F).
\end{equation}
For 2D registration tasks, paired 2D images $I^x$ and $I^y$ yield respective mask sets $\{M_k^{x}\}_{k=1}^{K^x}$ and $\{M_k^{y}\}_{k=1}^{K^y}$. For 3D registration tasks, as detailed in algorithm~\ref{algo: SAMReg}, paired 3D images $I^x$ and $I^y$, each sized $(H,W,D)$, are processed as a series of 2D slices $\{I^x_s\}_{s=1}^D$ and $\{I^y_s\}_{s=1}^D$ respectively. Each segmented ROI $R^x_k$ in $I^x_{s^\prime}$, corresponds to an ROI $R^y_k$ potentially located in any slice $\{I^y_s\}_{s=1}^D$. To optimize computation, we empirically limit our focus to a range $\delta s$ of slice $\{I^y_s\}_{s^\prime-\delta s}^{s^\prime+\delta s}$, sequentially processed to generate a comprehensive set $\{M_k^{y}\}_{k=1}^{K^y}$, which are then matched with $\{M_k^{x}\}_{k=1}^{K^x}$.

For each binary mask, we derive moving prototypes $\{p_k^{x}\}_{k=1}^{K^x}\subset \mathbb{R}^{1\times C^\prime}$ by element-wise multiplying the mask with their corresponding feature map, element being indexed by $n$: 
\begin{equation}
\label{eq:proto}
    p^x_k = \frac{\sum_{n}M^{x,res}_k(n)\cdot F^x(n)}{\sum_{n}M^{x,res}_k(n)},
\end{equation}
where $res$ indicates resizing $M^{x}_k\in \mathbb{R}^{H \times W}$ to the size of $F^x$ as $M^{x,res}_k \in \mathbb{R}^{H^\prime \times W^\prime}$. This also applies on the fixed prototypes $\{p_k^{y}\}_{k=1}^{K^y}\subset \mathbb{R}^{1\times C^\prime}$.

\textbf{\textit{\newnew{b) ROI Matching: }}} 
Assume a similarity matrix $S \in \mathbb{R}^{K^x \times K^y}$ to measure the cosine similarity between the moving- and fixed prototypes:
\begin{equation}
\label{eq:similarity}
    S(i,j)=\lVert\frac{p_i^{x}\cdot p_j^{y}}{\lVert p_i^{x} \rVert \cdot \lVert p_j^{y} \rVert}\rVert,
\end{equation}
 where $\lVert\cdot\rVert$ normalises each element $S(i,j)\in[0,1]$, $i\in[1,K^x]$ and $j\in[1,K^y]$.

A set of index pairs $P$ then can be identified: 
\begin{equation}\label{eq: pairing}
    P = \mathop{\arg\max}\limits_{i,j}(S(i,j)), \textit{s.t.}~ S(i,j)>\epsilon ~\forall~{(i,j)}, 
\end{equation}
corresponding to pairings with maximum similarities. A further constraint is added to ensure each $i, j$ is chosen at most once, if overlapping ROIs segmented from SAM are filtered. $\epsilon$ is the similarity threshold and the set length $K = |P| \leq min(K^x,K^y)$.

Using the selected index pairs $P$, we construct two new sets of masks, $\{M_k^{x,cor}\}_{k=1}^{K}$ and $\{M_k^{y,cor}\}_{k=1}^{K}$:
\begin{equation}
\label{eq:maskpairs}
  (M_k^{x,cor}, M_k^{y,cor}) = \{(M_i^{x}, M_j^{y}) | (i, j) \in P\}, k=1,2,...,K.  
\end{equation}

If an exact ``one-to-one'' mapping of correspondences is not strictly necessary, each $i,j$ may be selected repeatedly. The value of $K$ is established based on the threshold $\epsilon$.

\subsubsection{Optional post-SAMReg ROI-to-Dense correspondence conversion
}
\label{sec:dense}
Further, as discussed in Sec.~\ref{One registration}, we explore the capability of converting estimated ROI-based correspondence to its dense counterpart DDF, useful for applications such as full-image alignment. \newnew{It should be noted that this conversion is not part of the SAMReg algorithm, but serves as an optional post-processing step applied to the ROI pairs obtained by SAMReg when dense correspondence is needed.}
In this section, we describe a general method that iteratively refines the DDF, to optimize an objective function $\mathcal{L}$ that combines a region-specific alignment measure $\mathcal{L}_{roi}$ with a regularization term $\mathcal{L}_{ddf}$ to ensure smooth interpolation:

\begin{equation}
\mathcal{L}_{\Theta} = \textstyle \sum_{k=1}^K \mathcal{L}_{roi}(M_k^{x}, \mathcal{T}(M_k^{y},\Theta)) + \lambda \mathcal{L}_{ddf}(\Theta) 
\end{equation}

\newnew{where $\mathcal{T}$ denotes the transformation function that warps one set of paired ROI masks into the coordinate space of the other, $\Theta$ represents the parameters of $\mathcal{T}$, and $\lambda$ is a regularization parameter that balances the alignment accuracy with the deformation smoothness. }
This may itself be considered a multi-ROI alignment algorithm but with known ROI correspondence. In this work, we use an equally-weighted MSE and Dice as $\mathcal{L}_{roi}$ and a L$^2$-norm of DDF gradient as $\mathcal{L}_{ddf}$.

\begin{algorithm}
\caption{3D SAMReg Implementation}
\label{algo: SAMReg}
\begin{algorithmic}[1]
\State \textbf{Input:} Moving image $I^x\in\mathbb{R}^{H\times W \times D}$, fixed image $I^y\in\mathbb{R}^{H\times W \times D}$ and slice range $\delta s$
\State \textbf{Output:} Binary mask pairs $\{(M^x_k,M^y_k)\}_{k=1}^{K}$
\Procedure{RegisterImage}{$I^x,I^y$}
    \For{each slice $I^x_{s^\prime}$ in $I^x$}
        \State Derive ROI masks $\{M^x_k\}_{k=1}^{K^x}$ and corresponding prototypes $\{p^x_k\}_{k=1}^{K^x}$ with $I^x_{s^\prime}$ via Eq.~\ref{eq:encode},~\ref{eq:decode} and~\ref{eq:proto}
        \For{each slice $I^x_{s^{\prime\prime}}$ in range $(s^\prime-\delta s, s^\prime-\delta s)$ in $I^y_{s^{\prime\prime}}$}
            \State  Stack ROI masks $\{M^y_k\}_{k=1}^{K^y}$ and corresponding prototypes $\{p^y_k\}_{k=1}^{K^y}$ with $I^y_{s^{\prime\prime}}$ via Eq.~\ref{eq:encode},~\ref{eq:decode} and~\ref{eq:proto} as well
        \EndFor
        \State Compute $S\in \mathbb{R}^{K^x \times K^y}$ with prototypes via Eq.~\ref{eq:similarity}
        \State Establish index pairs $P$ with $S$ via Eq.~\ref{eq: pairing}
        \State Form mask pairs $\{(M^{x,cor}_k, M^{y,cor}_k)\}$ and assign to indices $P$ via Eq.~\ref{eq:maskpairs}
    \EndFor
    \State Append all pairs to the output list $\{(M^{x,cor}_k,M^{y,cor}_k)\}_{k=1}^{K}$
    \State \textbf{return} all paired masks $\{(M^{x,cor}_k,M^{y,cor}_k)\}_{k=1}^{K}$
\EndProcedure
\end{algorithmic}
\end{algorithm}

\begin{table*}
\footnotesize
\centering
\caption{Performance comparison with the-state-of-the-art (SOTA) medical image registration methods and SAMReg, a technique combined with different SAM-based segmentation algorithms on both 2D~\cite{LUNG,AERIAL} and 3D~\cite{bosaily2015promis,bernard2018deep,LUNG} datasets, Methods marked with an $*$ are specifically trained for the corresponding dataset. The best performance is in \textbf{bold}.}
\label{tab:sota}
\setlength{\tabcolsep}{1.5pt}
 \renewcommand{\arraystretch}{1.2} 
\begin{tabular}{c|cc|cc|cc|cc|cc} 
\hline
\multirow{3}{*}{\textbf{Method }} & \multicolumn{6}{c|}{\textbf{3-Dimensional Data}} & \multicolumn{4}{c}{\textbf{2-Dimensional Data}} \\ 
\cline{2-11}
 & \multicolumn{2}{c|}{\textbf{Prostate}} & \multicolumn{2}{c|}{\textbf{Cardiac}} & \multicolumn{2}{c|}{\textbf{Lung}} & \multicolumn{2}{c|}{\textbf{Retina}} & \multicolumn{2}{c}{\textbf{Aerial}} \\ 
\cline{2-11}
 & \textbf{Dice} & \textbf{TRE} & \textbf{Dice} & \textbf{TRE} & \textbf{Dice} & \textbf{TRE} & \textbf{Dice} & \textbf{TRE} & \textbf{Dice} & \textbf{TRE} \\ 
\hline
\textbf{NiftyReg}~\cite{modat2014global} & 7.68±3.98 & 4.67±3.48 & 9.93±2.21 & 3.29±2.89 & 10.93±2.02 & 4.23±1.64 & 8.77±3.29 & 5.53±3.52 & 10.21±3.02 & 4.02±2.25 \\
\textbf{VoxelMorph}$^*$~\cite{balakrishnan2019voxelmorph} & 56.84±3.41 & 3.68±1.92 & 60.10±3.95 & 3.03±2.41 & 77.98±2.72 & 3.23±0.81 & 53.12±3.46 & 4.66±3.43 & 72.73±2.41 & 3.53±1.3 \\
\textbf{LabelReg}$^*$~\cite{hu2018weakly} & 77.32±3.56 & 2.72±1.23 & 78.97±2.42 & 1.73±1.34 & 83.56±2.43 & 1.52±0.86 & 69.73±3.51 & 2.69±3.21 & 83.35±2.14 & 3.51±1.47 \\ 
\hline
\multicolumn{1}{l|}{\textbf{SAMReg~}\textit{w} \scriptsize SAM~\cite{kirillov2023segment}} & \textbf{75.67±3.81} & \textbf{2.09±1.35} & \textbf{80.28±3.67} & \textbf{1.43±1.13} & 85.23±2.16 & 1.31±0.91 & 61.79±3.42 & 3.44±3.35 & 84.13±2.23 & 3.46±1.61 \\
\multicolumn{1}{l|}{\textbf{SAMReg~}\textit{w} \scriptsize MedSAM~\cite{ma2024segment}} & 70.32±3.52 & 2.11±1.66 & 78.66±3.71 & 1.59±1.21 & \textbf{86.34±2.12} & \textbf{1.27±0.83} & \textbf{64.33±3.38} & 2.89±3.27 & 82.19±2.56 & 3.99±1.52 \\
\multicolumn{1}{l|}{\textbf{SAMReg~}\textit{w} \scriptsize SAMed2D~\cite{cheng2023sammed2d}} & 71.23±3.77 & 2.49±1.17 & 77.52±3.75 & 1.65±1.29 & 86.16±2.15 & 1.27±0.86 & 63.10±3.35 & \textbf{2.81±3.22} & 80.72±2.98 & 3.96±1.76 \\
\multicolumn{1}{l|}{\textbf{SAMReg~}\textit{w} \scriptsize SAMSlim~\cite{chen20230}} & 70.66±3.79 & 2.48±1.24 & 78.23±3.67 & 1.57±1.20 & 84.53±2.34 & 1.42±0.90 & 60.53±3.46 & 3.40±3.31 & 83.52±2.20 & 3.49±1.62 \\
\multicolumn{1}{l|}{\textbf{SAMReg~}\textit{w} \scriptsize SAMHQ~\cite{ke2024segment}} & 73.83±3.69 & 2.26±1.15 & 79.11±3.62 & 1.48±1.15 & 84.78±2.29 & 1.39±0.89 & 61.87±3.39 & 3.36±3.28 & \textbf{85.17±2.18} & \textbf{3.36±1.59} \\
\hline
\end{tabular}
\end{table*}

\begin{figure*}[!t]
    \centering
    \includegraphics[width=0.7\textwidth]{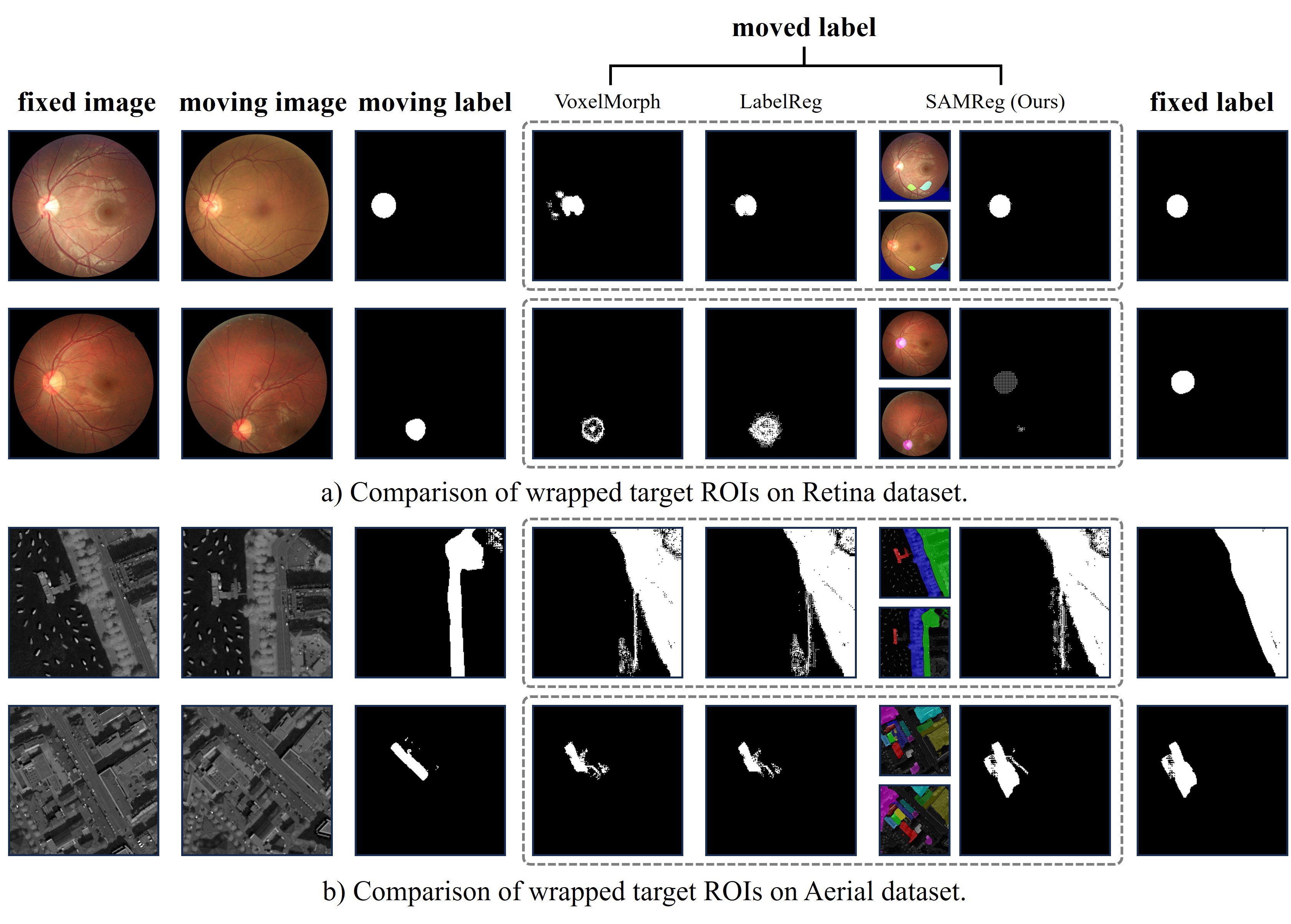}
    \caption{Qualitative comparisons of SAMReg with other leading registration methods~\cite{balakrishnan2019voxelmorph,hu2018weakly} on 2D datasets: a) Retina and b) Aerial. Generated DDF aligns the ROI from the moving to the moved label, matching the fixed label's ROI. Colored areas in the SAMReg column highlight the corresponding ROIs, which closely resemble the fixed label, showcasing superior alignment performance.}
    \label{fig:2Dqualitative}
\end{figure*}
\begin{figure*}[!t]
    \centering
    \includegraphics[width=0.7\textwidth]{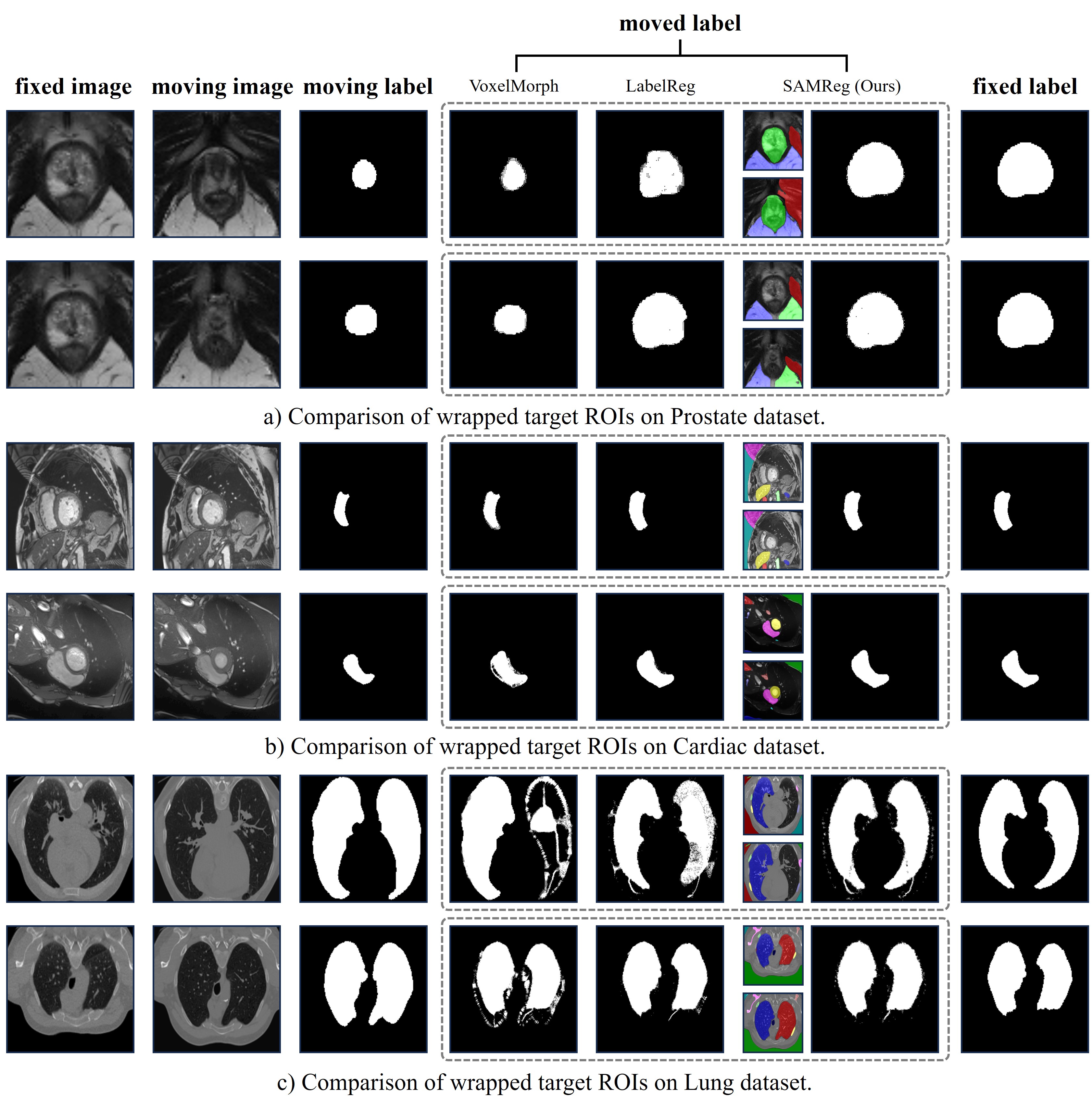}
    \caption{Qualitative comparisons of SAMReg with other leading registration methods~\cite{balakrishnan2019voxelmorph,hu2018weakly} on 3D datasets: a) Prostate~\cite{bosaily2015promis}, b) Cardiac~\cite{bernard2018deep} and c) Lung~\cite{LUNG}. Generated DDF maps ROIs from the moving to the moved label, aligning with the fixed label's ROIs. Colored highlights in the SAMReg column indicate the matched ROIs by SAMReg, demonstrating its superior registration performance.}
    \label{fig:3Dqualitative}
\end{figure*}

\section{Experiments}
\label{sec:exp_results}

\subsection{Datasets}

\new{We evaluate our method in five medical and non-medical datasets (Table~\ref{tab:dataset})}: Prostate~\cite{bosaily2015promis,hamid2019smarttarget,orczyk2021prostate,dickinson2013multi}, Cardiac~\cite{bernard2018deep}, Lung~\cite{LUNG}, Retina~\cite{HISTOLOGY} and Aerial~\cite{AERIAL}. The first three datasets are three-dimensional and contain MR and CT imaging modalities, while the latter two are two-dimensional. \textbf{Prostate}: Consisting of $878$ T2-weighted MR images derived from several clinical trials at University College London Hospital. \textbf{Cardiac}: Consisting of $150$ exams acquired at the University Hospital of Dijon and proposed in Automated Cardiac Diagnosis Challenge (ACDC) challenge~\cite{bernard2018deep}. Each clinical exams contains pair MR cardiac images from different frames. \textbf{CT-Lung}: Consisting of $20$ pairs of FBCT images from the Learn2Reg 2021 challenge~\cite{LUNG}, this dataset provides images capturing both expiratory and inspiratory states. \textbf{Retina}: comprising $134$ retinal image pairs from Fundus Image Registration Dataset (FIRE)~\cite{hernandez2017fire}, acquired at the Papageorgiou Hospital, Aristotle University of Thessaloniki, Thessaloniki from $39$ patients. \textbf{2D-Aerial}: comprising $1734$ pairs of QuickBird-acquired images of the city of Zurich~\cite{AERIAL}. Image side lengths range from $622$ to $1830 ~px$. 

Further, to ensure a fair comparison with other supervised registration methods, 80\% of the images are utilized for training purposes. Our method does not use this subset for training and is tested on the remaining data alongside other methods. For ablative experiments, the proposed algorithms are tested using all available images. 
\begin{table}
\centering
\caption{Ablation study of segmentation performance: SAM versus the respective state-of-the-art (SOTA) dataset-specific segmentation models, on 3D datasets~\cite{bosaily2015promis,bernard2018deep,LUNG}. Both label ROIs (ground truth) and pseudo ROIs identified by SAM are evaluated.}
\label{tab:seg_model}
\footnotesize
\setlength{\tabcolsep}{1.5pt}
 \renewcommand{\arraystretch}{1.2} 
\begin{tabular}{c|c|c|c|c|c} 
\hline
\multirow{2}{*}{\textbf{Dataset}} & \multirow{2}{*}{\begin{tabular}[c]{@{}c@{}}\textbf{Segmentation}\\\textbf{Algorithm}\end{tabular}} & \multicolumn{2}{c|}{\textbf{Label ROI}} & \multicolumn{2}{c}{\textbf{Pseudo ROI}} \\ 
\cline{3-6}
 &  & \textbf{Dice} & \textbf{TRE} & \textbf{Dice} & \textbf{TRE} \\ 
\hline
\multirow{2}{*}{\textbf{Prostate}~\cite{bosaily2015promis}} & \textbf{SAM}~\cite{kirillov2023segment} & 75.67±3.81 & 2.72±1.23 & 97.73±3.15 & 0.82±0.23 \\
 & \textbf{Vnet}~\cite{milletari2016v} & 80.35±3.14 & 2.13±1.24 & 43.56±3.54 & 5.32±2.51 \\ 
\hline
\multirow{2}{*}{\textbf{Cardiac}~\cite{bernard2018deep}} & \textbf{SAM}~\cite{kirillov2023segment} & 80.28±3.67 & 1.73±1.34 & 98.87±3.61 & 1.62±0.67 \\
 & \textbf{FCT}~\cite{tragakis2023fully} & 83.54±3.46 & 1.31±1.25 & 31.35±3.74 & 7.24±2.52 \\ 
\hline
\multirow{2}{*}{\textbf{Lung}~\cite{LUNG}} & \textbf{SAM}~\cite{kirillov2023segment} & 85.23±2.16 & 1.31±0.91 & 98.94±3.24 & 1.31±0.63 \\
 & \textbf{MedNeXt}~\cite{roy2023mednext} & 89.78±3.19 & 1.10±1.45 & 50.16±3.27 & 5.01±2.33 \\
\hline
\end{tabular}
\end{table}
\subsection{Implementation Details}

For non-rigid registration~\cite{eppenhof2018deformable,sotiras2013deformable}, we assess accuracy by the mean transformation accuracy of critical anatomical structures~\cite{hu2018weakly}, with TRE calculated from the centroids of the relevant ROIs. \newnewnew{For Dice evaluation, the estimated transformation is applied to the segmentation label. }
In configuring the SAM parameters, beyond the standard settings, we adjusted the pred-iou-thresh and stability-score-thresh to 0.90. Our ROI filtering strategy is based on area and overlapping ratio, \new{setting a minimum of 200 and a maximum of 7000 for the ROI sizes and a maximum ratio of 0.8. The similarity threshold $\epsilon$ is set to 0.8.} \newnew{For most applications, we recommend using the default settings reported in this manuscript. Nevertheless, SAMReg allows these inference-stage parameters to be adjusted flexibly without retraining. For example, $\delta s$ can be increased when larger through-plane displacement or lower through-plane resolution is expected, while $\epsilon$ can be adjusted according to the desired similarity threshold for ROI matching.} \new{Our network is implemented with PyTorch and MONAI~\cite{cardoso2022monai}. }

\section{Results}

\subsection{Comparison with SOTA Registration Methods}

\new{In our comprehensive evaluation (Table~\ref{tab:sota}), we present comparisons between the proposed SAMReg's performance against both non-learning and learning-based medical image registration techniques on both 2D and 3D datasets. SAMReg is fairly compared with NiftyReg~\cite{modat2014global}, a well-established registration tool for non-learning iterative algorithms, without using any training data for both. It is also compared with adapted methods including VoxelMorph~\cite{balakrishnan2019voxelmorph} and LabelReg~\cite{hu2018weakly}, although these methods require training data. VoxelMorph and LabelReg are fine-tuned on the specific datasets, employing unsupervised and weakly-supervised learning respectively, with LabelReg also depending on anatomical annotations. 
SAMReg achieves competitive performance, particularly in intra-subject cardiac and lung registrations. 
While Dice in inter-subject prostate registrations might be marginally lower than the learning-based methods, SAMReg's competitive TRE suggests a focus on achieving high local alignment accuracy. This might be due to the inherently larger anatomical variations present in the prostate glands compared to organs like the hearts or lungs. 
Furthermore, SAMReg performs well on non-medical tasks like aerial image registration, indicating its domain-independent salient image feature matching and broad applicability to diverse computer vision and medical imaging scenarios.}

\textcolor{black}{Qualitative comparisons of 2D and 3D registration results are shown in Figs.~\ref{fig:2Dqualitative} and~\ref{fig:3Dqualitative}, respectively.} 
\textcolor{black}{Compared to other registration methods, DDF derived from multiple ROI pairs enables accurate warping of the moving label to closely match the fixed label, even if the ROI is unseen by SAMReg.}

\subsection{Comparison of SOTA ROI Segmentation Algorithms}
\textcolor{black}{SAMReg's performance can be further enhanced through the direct integration of advanced segmentation algorithms. This can be implemented in two ways: 1) dataset-specific integration - for specific datasets and tasks, SAMReg can be efficiently combined with other pre-trained segmentation models optimized for that domain. 2) general-purpose integration - for broader applicability, SAMReg can be effectively utilized with SAM-based segmentation, offering a generic solution for various registration tasks.}

\subsubsection{Impact of Dataset-Specific SOTA Segmentation Models}
\begin{figure}[!t]
    \centering
    \includegraphics[width=0.5\textwidth]{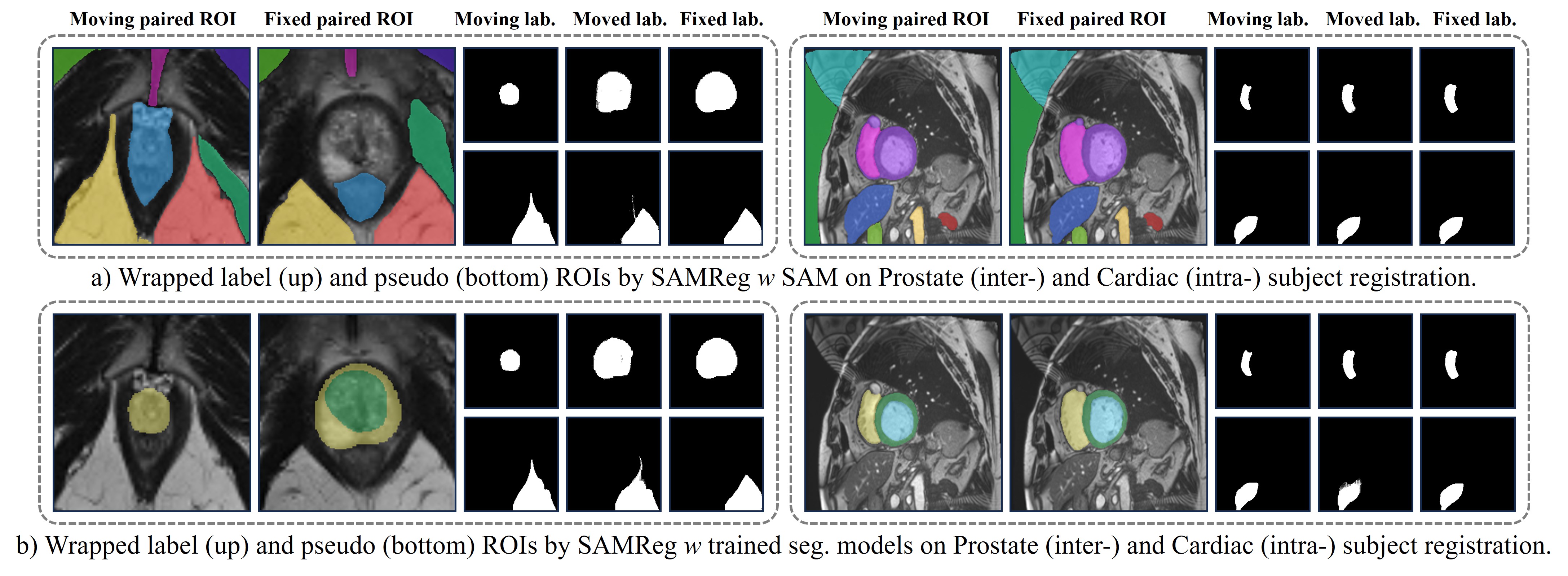}
    \caption{The qualitative comparisons of a) SAM~\cite{kirillov2023segment} and b) fully-supervised trained segmentation models~\cite{milletari2016v,tragakis2023fully,roy2023mednext} on inter- and intra-subject registration tasks, including the wrapping performance on label ROI (top) and pseudo ROI (bottom). SAMReg with SAM benefits from its extensive insight. Despite being fully trained, dataset-specific segmentation models show competitive performance only on familiar ROIs (label ROI) but falter on unknown ROIs (pseudo ROI).  }
    \label{fig:trained_model}
\end{figure}

\textcolor{black}{Table~\ref{tab:seg_model} and Fig.~\ref{fig:trained_model} present a comparison of ROI segmentation and subsequent registration alignment performance using SAM~\cite{kirillov2023segment} compared to leading segmentation models in specific datasets~\cite{milletari2016v,tragakis2023fully,roy2023mednext}.} \newnewnew{The SAM results are repeated from Table II as a common reference for direct comparison.} \textcolor{black}{While the benchmark models benefit from being specifically designed} and trained on ground-truth annotations (label ROI), SAM-based SAMReg demonstrates robustness and effectiveness in registration tasks. Notably, SAM achieves competitive performance, even surpassing these models, when utilizing SAM-generated pseudo ROIs (pseudo ROI) instead of ground-truth data. Furthermore, SAM's extensive ROI detection capabilities demonstrably enhance image analysis and strengthen the overall registration accuracy. This is particularly evident when dealing with previously unseen ROIs, highlighting SAM-based SAMReg's interesting advantage in handling both predefined and unknown ROIs effectively.

\begin{figure}[!t]
    \centering
    \includegraphics[width=0.5\textwidth]{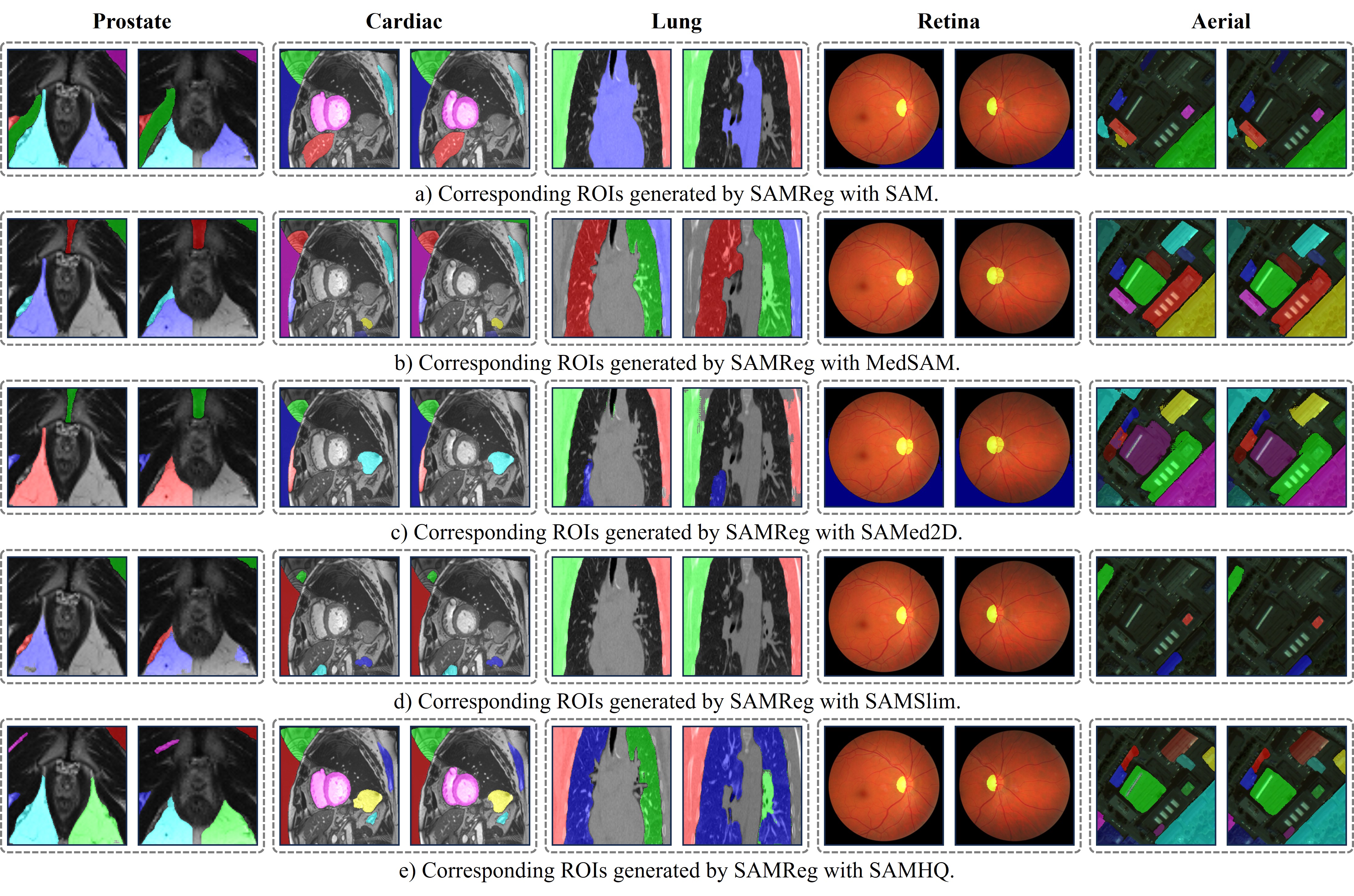}
    \caption{\new{The qualitative comparisons of SAMReg paired with various SAM-based segmentation models: a) SAM, b) MedSAM, c) SAMed2D, d) SAMSlim, and e) SAMHQ, across multiple datasets including Prostate, Cardiac, Lung, Retina, and Aerial.} MedSAM and SAMed2D are specialized medical models, whereas SAM, SAMSlim, and SAMHQ serve general purposes. Each configuration generates unique paired ROIs, demonstrating their respective registration performance.}
    \vspace{-1.5em}
    \label{fig:SAM_type}
\end{figure}

\subsubsection{Impact of SAM-Based SOTA Segmentation Models}

Table~\ref{tab:sota} compares the SAMReg performance with SAM~\cite{kirillov2023segment} and its variants when applied to medical and general registration tasks. Specifically, MedSAM~\cite{ma2024segment} and SAMed2D~\cite{cheng2023sammed2d} are fine-tuned on large medical datasets for medical fundamental segmentation, while SAMSlim~\cite{chen20230} and SAM\_HQ~\cite{ke2024segment} are modified for enhancing general segmentation performance. Examples of the segmented and paired ROIs of these SAM variants on all five datasets are visualized in Fig.~\ref{fig:SAM_type}.

The results demonstrate that medical SAM variants achieve more accurate anatomical structure segmentation compared to general SAM variants, leading to better alignment performance. Conversely, general SAM variants outperform medical SAM variants in non-medical registration tasks, \textit{i.e.}, aerial dataset. Interestingly, the basic SAM consistently achieves competitive results across all tasks, which may be due to the fact that it generates more paired ROIs as illustrated in Fig.~\ref{fig:SAM_type}.
\begin{table}
    \caption{Ablation study of paired ROI quantity threshold on intra- (cardiac~\cite{bernard2018deep}) and inter-subject (prostate~\cite{bosaily2015promis}) registration tasks. The best performance is in bold.}
\label{tab:abla_t_threshold}
\centering
\footnotesize
\setlength{\tabcolsep}{1.5pt}
 \renewcommand{\arraystretch}{1.2} 
\begin{tabular}{c|cc|cc} 
\hline
\multirow{2}{*}{\begin{tabular}[c]{@{}c@{}}\textbf{Quantity}\\\textbf{Threshold~}\end{tabular}} & \multicolumn{2}{c|}{\textbf{Intra-Subject: Cardiac}} & \multicolumn{2}{c}{\textbf{Inter-Subject: Prostate}} \\ 
\cline{2-5}
 & \textbf{Dice} & \textbf{TRE} & \textbf{Dice} & \textbf{TRE} \\ 
\hline
\textbf{1} & 73.31±3.87 & 1.73±1.34 & 69.59±3.51 & 3.85±1.59 \\
\textbf{3} & 78.26±3.77 & 1.58±1.21 & \textbf{76.00±3.23} & \textbf{2.06±1.37} \\
\textbf{5} & \textbf{79.45±3.72} & \textbf{1.47±1.18} & 75.36±3.78 & 2.11±1.39 \\
\textbf{7} & 78.38±3.76 & 1.49±1.20 & 74.75±3.80 & 2.15±1.37 \\
\textbf{9} & 77.31±3.79 & 1.52±1.25 & 72.89±3.83 & 2.22±1.41 \\
\hline
\end{tabular}
\end{table}
\begin{table}
   \caption{Ablation study of paired ROI similarity threshold $\epsilon$  on intra- (cardiac~\cite{LUNG}) and inter-subject (prostate~\cite{bosaily2015promis}) registration tasks. The best performance is in bold.}
\label{tab:abla_q_threshold}
\centering
\footnotesize
\setlength{\tabcolsep}{1.5pt}
 \renewcommand{\arraystretch}{1.2} 
\begin{tabular}{c|cc|cc} 
\hline
\multirow{2}{*}{\begin{tabular}[c]{@{}c@{}}\textbf{Similarity}\\\textbf{Threshold}\end{tabular}} & \multicolumn{2}{c|}{\textbf{Intra-Subject: Cardiac}} & \multicolumn{2}{c}{\textbf{Inter-Subject: Prostate}} \\ 
\cline{2-5}
 & \textbf{Dice} & \textbf{TRE} & \textbf{Dice} & \textbf{TRE} \\ 
\hline
\textbf{0.7} & 78.69±3.73 & 1.70±1.29 & 74.41±3.91 & 2.19±1.39 \\
\textbf{0.75} & 80.21±3.69 & 1.54±1.16 & 75.13±3.87 & 2.13±1.36 \\
\textbf{0.8} & \textbf{80.28±3.67} & \textbf{1.43±1.13} & \textbf{75.67±3.81} & \textbf{2.09±1.35} \\
\textbf{0.85} & 79.11±3.70 & 1.65±1.20 & 75.45±3.79 & 2.11±1.33 \\
\textbf{0.9} & 77.93±3.72 & 1.72±1.23 & 74.12±3.84 & 2.12±1.34 \\
\textbf{0.95} & 77.12±3.77 & 1.73±1.21 & 73.80±3.86 & 2.17±1.38 \\
\hline
\end{tabular}
\end{table}

\begin{figure*}[!t]
    \centering
    \includegraphics[width=0.75\textwidth]{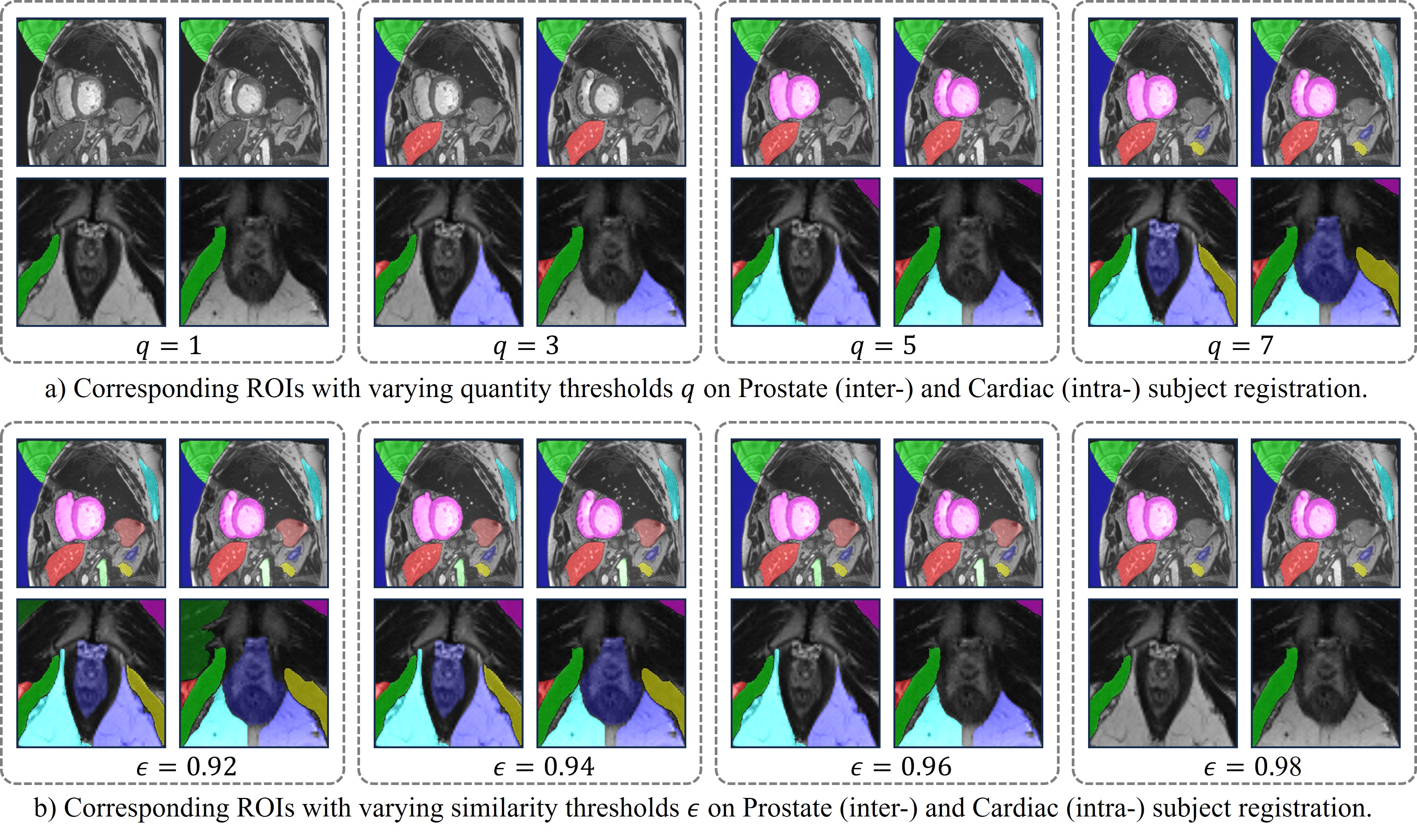}
    \caption{The qualitative comparisons of SAMReg set as various thresholds: a) quantity $K$ and b) similarity $\epsilon$, on inter- and intra-subject registration tasks. Increasing $K$ introduces mismatches and noise as $\epsilon$ decreases; raising $\epsilon$ limits the field of view and reduces $K$.}
    \vspace{-1em}
    \label{fig:abla_threshold}
\end{figure*}

\begin{table}
\footnotesize
\centering
\caption{Ablation study of slice range $\delta s$ on intra- (cardiac~\cite{bernard2018deep}) and inter-subject (prostate~\cite{bosaily2015promis}) registration tasks. The best performance is in bold.}
\label{tab:abla_slice}
\setlength{\tabcolsep}{1.5pt}
 \renewcommand{\arraystretch}{1.2} 
\begin{tabular}{c|cc|cc} 
\hline
\multirow{2}{*}{\begin{tabular}[c]{@{}c@{}}\textbf{Slice}\\\textbf{Range (±)}\end{tabular}} & \multicolumn{2}{c|}{\textbf{Intra-Subject: Cardiac}} & \multicolumn{2}{c}{\textbf{Inter-Subject: Prostate}} \\ 
\cline{2-5}
 & \textbf{Dice} & \textbf{TRE} & \textbf{Dice} & \textbf{TRE} \\ 
\hline
\textbf{1} & 71.23±3.54 & 2.16±1.34 & 61.29±4.05 & 2.77±1.53 \\
\textbf{3} & 73.58±3.59 & 1.84±1.26 & 65.12±4.01 & 2.65±1.59 \\
\textbf{5} & 76.37±3.44 & 1.67±1.21 & 68.59±3.97 & 2.52±1.52 \\
\textbf{7} & 78.55±3.51 & 1.58±1.23 & 71.66±3.92 & 2.28±1.45 \\
\textbf{9} & 79.13±3.62 & 1.49±1.19 & 73.54±3.91 & 2.15±1.39 \\
\textbf{11} & \textbf{80.28±3.67} & \textbf{1.43±1.13} & \textbf{75.67±3.81} & \textbf{2.09±1.35} \\
\hline
\end{tabular}
\end{table}

\subsection{Ablative Study on ROI Selecting Strategies}
In this section, we conduct an ablative study on ROI selecting strategies, which define the correspondence and the subsequent generated DDFs, testing different threshold values in similarity or quantity, as described in Sec.~\ref{sec:align}.

\subsubsection{Impact of Quantity Threshold}
The impact of the number of selected ROI pairs on registration performance is evaluated. \newnewnew{Table~\ref{tab:abla_t_threshold}} shows Dice and TRE scores for different quantities of ROI pairs on intra-subject (cardiac) and inter-subject (prostate) datasets. The results suggest that using up to \newnewnew{three ROI pairs} achieves the best registration performance on the prostate dataset and \newnewnew{five} on cardiac dataset, suggesting that a relatively small number of well-chosen ROIs can be sufficient for achieving optimal registration. As shown in Fig.~\ref{fig:abla_threshold}a,  additional paired ROIs improve the entire image alignment but introduce inherent noise due to segmentation model limitations.

\subsubsection{Impact of Similarity Threshold}

We further investigate the effect of a similarity threshold on ROI selection. \newnewnew{Table~\ref{tab:abla_q_threshold}} illustrates the Dice and TRE scores for different similarity thresholds on intra-subject (cardiac) and inter-subject (prostate) datasets. The results demonstrate that a 0.8 similarity threshold leads to the best overall registration performance, indicating that using ROIs with a moderately high degree of similarity ($\leq$ 0.8) between the source and target images is beneficial for registration. A higher threshold ($\geq$ 0.8) reduces the number of selected ROIs, potentially leading to a lack of sufficient corresponding information for accurate registration, reflected in the higher TRE scores. Conversely, a lower threshold might incorporate too many ROIs with potentially irrelevant similarities, as illustrated in Fig.~\ref{fig:abla_threshold}b, hindering the registration performance. These findings underscore the importance of striking a balance between the similarity and quality threshold of selected paired ROIs for achieving optimal registration results.

As analyzed in Sec.~\ref{sec:roi-correspondence}, increasing the number or precision of ROI pairs generally improves performance up to a task-specific upper limit. However, as shown in Tables~\ref{tab:abla_q_threshold} and~\ref{tab:abla_t_threshold}, there is an optimal balance between the number/precision of ROI pairs and registration performance, which falls below the theoretical maximum. Beyond this balance point, further increasing the number or precision of ROI pairs results in a performance decline.

This counter-intuitive phenomenon occurs because segmentation and alignment performance are assumed constant in Sec.~\ref{sec:roi-correspondence}. In SAMReg’s ROI matching mechanism, ROI pairs are ranked by similarity, with the most similar pairs ranked highest. Adding more pairs reduces the average similarity, introducing noise and potential mismatches. Conversely, stricter thresholds reduce the number of pairs, limiting the benefits of ROI-to-dense interpolation. Future work should explore adaptive thresholding strategies that dynamically adjust based on image similarity or segmentation confidence to optimize the number of pairs while maintaining high-quality correspondences.

\begin{table}
\footnotesize
\centering
\caption{Ablation study of ROI mapping: ``one-to-one'' versus ``one-to-many'' correspondence, on intra- (cardiac~\cite{bernard2018deep}) and inter-subject (prostate~\cite{bosaily2015promis}) registration tasks. }
\label{tab:abla_121}
\setlength{\tabcolsep}{1.5pt}
 \renewcommand{\arraystretch}{1.2} 
\begin{tabular}{c|cc|cc} 
\hline
\multirow{2}{*}{\textbf{Correspondence}} & \multicolumn{2}{c|}{\textbf{Intra-Subject: Cardiac}} & \multicolumn{2}{c}{\textbf{Inter-Subject: Prostate}} \\ 
\cline{2-5}
 & \textbf{Dice} & \textbf{TRE} & \textbf{Dice} & \textbf{TRE} \\ 
\hline
\textbf{``one-to-one''} & 80.28±3.67 & 1.43±1.13 & 75.67±3.81 & 2.09±1.35 \\
\textbf{``one-to-many''} & 78.12±3.89 & 1.65±1.19 & 75.12±4.03 & 2.03±1.62 \\
\hline
\end{tabular}
\vspace{-1em}
\end{table}

\begin{figure}[htbp]
    \centering
    \includegraphics[width=0.3\textwidth]{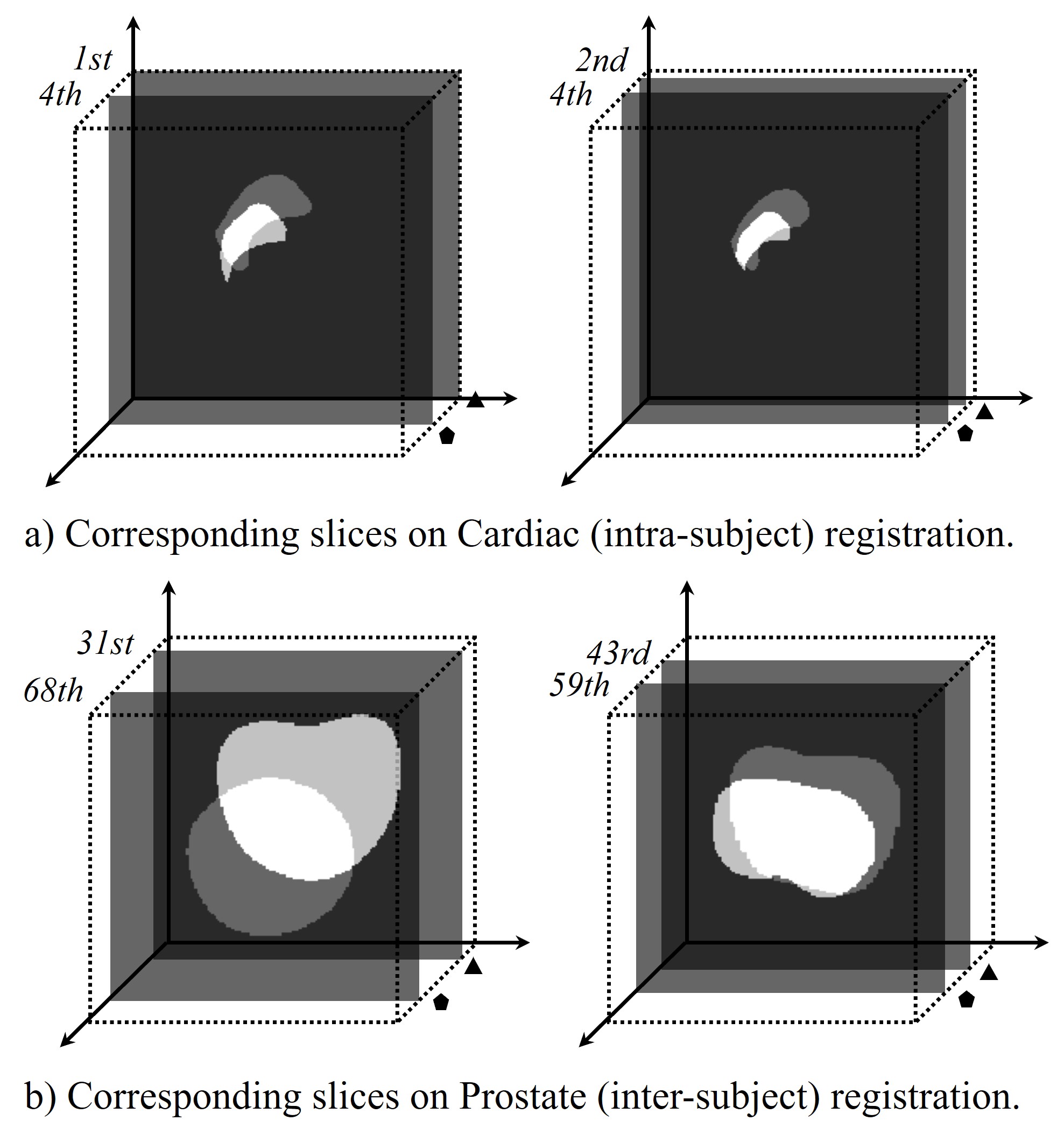}
    \caption{Examples of corresponding ROIs, marked as black pentagons and triangles in the bottom right, are shown across various slices on inter-subject registration tasks.}
    \vspace{-1em}
    \label{fig:abla_slice}
\end{figure}
\subsection{Ablative Study on ROI Mapping Strategies}
\label{sec: ROI mapping}
\subsubsection{Comparison Between `One-to-One' and `One-to-Many' Correspondence}
The inherent uncertainty in segmented ROIs necessitates exploring appropriate correspondence representation, such as ``one-to-one'' matching or allowing ``one-to-many'' relationships. In this study, ``one-to-one'' correspondence pairs each ROI in one image with a unique ROI in the other, while ``one-to-many'' correspondence allows for multiple ROIs from one image to correspond to a single ROI in the other.

Two key scenarios necessitate considering ``one-to-many'' correspondence: 1)
\textbf{Segmentation Uncertainty}: Segmentation models are not perfect. They might segment multiple sub-structures in one image while identifying a complete structure ROI in another. ``one-to-many'' correspondence can bridge this gap by allowing multiple sub-structures to correspond to a single complete structure. 2) \textbf{2D ROIs for 3D Structures}: Representing 3D structures with 2D ROIs exists inherent limitations. For instance, a single gland might be represented by 20 2D slices in one image and 40 in the other. This disparity requires ``one-to-many'' correspondence to handle multiple-to-one correspondence scenarios. 

As evidenced by Table~\ref{tab:abla_121}, the performance of both ``one-to-one'' and ``one-to-many'' approaches is comparable in intra-subject registration tasks, where images exhibit high similarity. \new{However, for inter-subject registration tasks with greater anatomical variability, while 'one-to-one' maintains a higher Dice, the 'one-to-many' strategy achieves a superior TRE. This experimental observation indicates that when processing complex anatomical deformations or cross-dimensional projections, relaxing the bijection constraint facilitates the capture of dispersed anatomical features.}

\subsubsection{Effect of Slice Range $\delta s$}
\new{Table~\ref{tab:abla_slice} presents the effect of the slice search range $\delta s$ on 3D registration performance for both intra-subject (cardiac) and inter-subject (prostate) tasks. A larger search range consistently improves registration accuracy, with both Dice and TRE improving as $\delta s$ increases from $\pm 1$ to $\pm 11$, although the rate of improvement begins to saturate as it approaches 11. This initial improvement is expected, as a wider search range permits matching 2D ROIs across a broader set of candidate slices, increasing the likelihood of identifying correct cross-slice correspondences for 3D structures.}

\new{Fig.~\ref{fig:abla_slice} illustrates examples of corresponding ROIs identified across different slices, demonstrating how the search range influences the spatial extent of matched structures in inter-subject registration tasks.}

\section{Conclusion}
In this study, we introduce a novel ROI-based correspondence representation. With this representation, the image registration is reformalized as two multi-class segmentation tasks, with a proposed, general-purpose and practical implementation, SAMReg. Our extensive experiments demonstrate that SAMReg delivers competitive performance in image registration, positioning it as a robust and versatile tool. The promising results suggest that this new approach may pave the way for a fresh research direction in image registration. 
\new{Despite these promising results, several aspects warrant further exploration to facilitate broader clinical translation. First, while evaluated across multiple datasets, future large-scale, multi-center studies will be beneficial to fully establish its statistical robustness. Second, because SAMReg builds upon a foundation model trained primarily on natural images, its performance and potential domain bias in medical modalities with highly ambiguous boundaries could be further improved. Third, while the current method relies on a generic spatial smoothness penalty, integrating explicit biomechanical priors could enhance its applicability in specific clinical workflows that require strictly physically plausible deformations. Future work could also incorporate explicit geometric descriptors as auxiliary constraints to improve spatial consistency in complex inter-subject deformations.}

\bibliographystyle{ieeetr}
\bibliography{refs}

\end{document}


\title{SAMReg: SAM-enabled Image Registration with ROI-based Correspondence}
\author{Shiqi Huang, Tingfa Xu, Ziyi Shen, Shaheer U Saeed, Wen Yan, Dean Barratt, Yipeng Hu }
\maketitle

\section{Supplementary Material}

\begin{figure*}[!t]
    \centering
    \includegraphics[width=\linewidth]{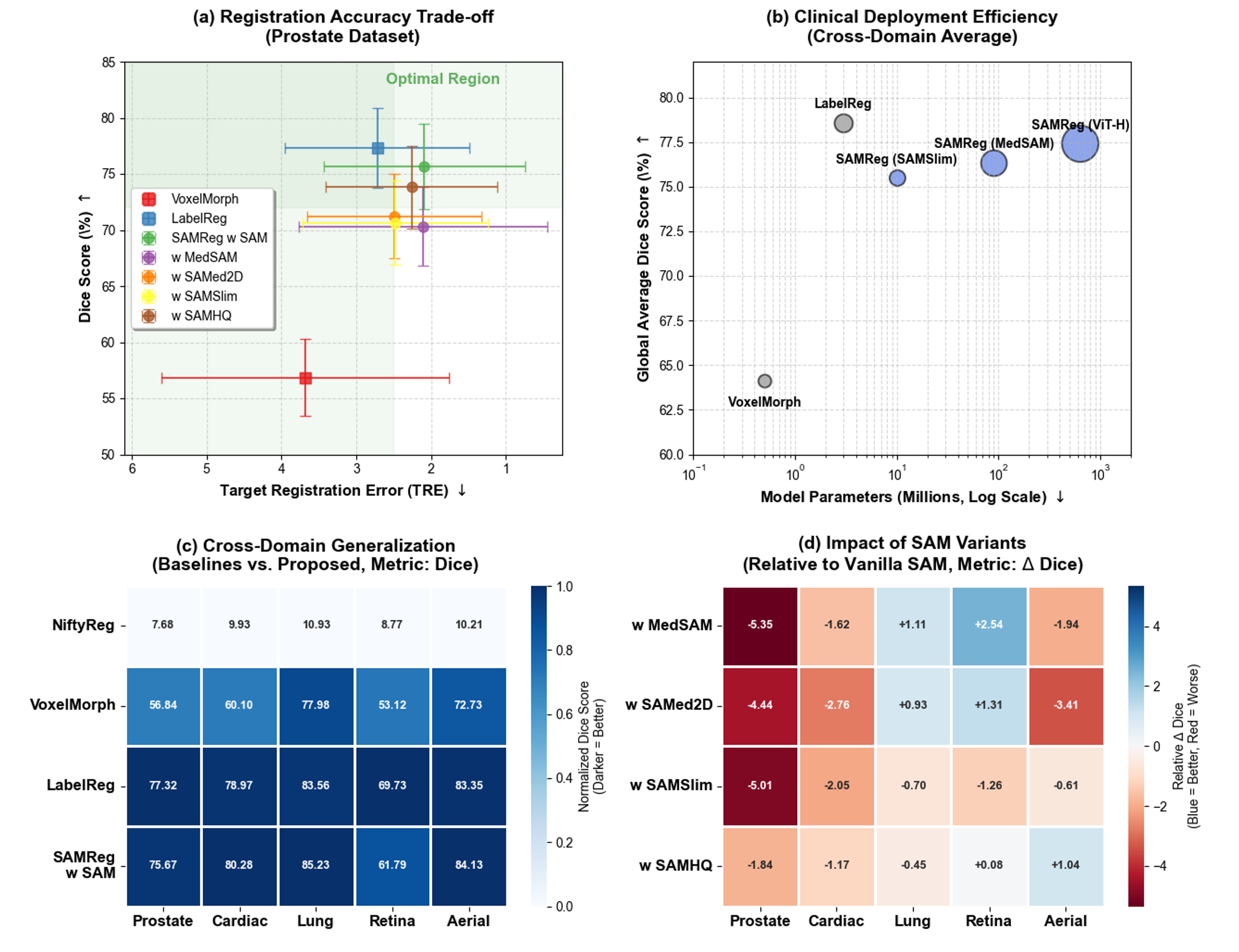}
    \caption{Summary of quantitative registration results from four analytical perspectives.
    (a)~Dice vs.\ TRE on the Prostate dataset; the green region indicates high Dice and low TRE. Error bars show standard deviations.
    (b)~Average Dice (five domains) vs.\ model parameter count (log-scale); bubble size reflects the parameter count.
    (c)~Column-normalized Dice heatmap (0--1 per dataset) comparing SAMReg with baseline methods across all five datasets.
    (d)~Relative Dice difference ($\Delta$Dice) when replacing the vanilla SAM backbone with medical-specific or lightweight variants; blue denotes improvement and red denotes degradation.}
    \label{fig:heatmap}
\end{figure*}

Fig.~\ref{fig:heatmap} provides a multi-perspective summary of the quantitative registration results, complementing the main-text tables by visualizing accuracy, efficiency, generalization, and the effect of SAM backbone variants.

\textbf{Registration accuracy} (Fig.~\ref{fig:heatmap}a). On the Prostate dataset, SAMReg variants cluster in the high-Dice, low-TRE region (green shading), indicating favourable trade-offs between region overlap and landmark alignment compared with learning-based baselines.

\textbf{Model efficiency} (Fig.~\ref{fig:heatmap}b). The average Dice across all five domains is plotted against the number of model parameters (log-scale). Among SAMReg configurations, lighter backbones such as SAMSlim offer a practical balance between performance and computational cost, whereas the full ViT-H backbone yields the highest accuracy.

\textbf{Cross-domain generalization} (Fig.~\ref{fig:heatmap}c). A column-normalized heatmap of Dice scores (scaled 0--1 within each dataset) shows that SAMReg maintains consistently high performance across all five domains, including both 2D/3D and medical/non-medical tasks, without domain-specific training.

\textbf{Impact of SAM variants} (Fig.~\ref{fig:heatmap}d). A relative-difference heatmap ($\Delta$Dice) compares each SAM variant against the vanilla SAM baseline (white). Blue and red cells denote performance gains and drops, respectively. The results indicate that medical-specific variants (e.g., MedSAM) tend to improve Dice on clinical datasets, while the vanilla SAM provides the most stable overall performance owing to its denser and more general feature representations.

